\documentclass[conference]{IEEEtran}
\IEEEoverridecommandlockouts
\usepackage{hyperref}
\usepackage{cite}
\usepackage{amsmath,amssymb,amsfonts}
\usepackage{algorithmic}
\usepackage{graphicx}
\usepackage{textcomp}
\usepackage{cleveref}
\usepackage{xcolor}
\usepackage{stfloats} 
\usepackage{booktabs} 
\usepackage{tabularx}        
\usepackage{makecell}        
\usepackage{float}   
\usepackage{xspace}
\usepackage{caption}
\usepackage[utf8]{inputenc} 
\usepackage{array}
\usepackage{placeins}
\begin{document}

\title{Recognition of Abnormal Events in Surveillance Videos using Weakly Supervised Dual-Encoder Models
\\
\thanks{*These authors contributed equally to this work.}
}

\author{
\makebox[\textwidth][c]{%
\begin{minipage}[t]{0.48\textwidth}\centering
Noam Tsfaty*\\
\textit{Intelligence Systems}\\
\textit{Afeka College of Engineering}\\
Tel Aviv, Israel\\
Email: Tsfaty.Noam@s.afeka.ac.il\\
ORCID: 0009-0009-5246-8274
\end{minipage}
\hfill
\begin{minipage}[t]{0.48\textwidth}\centering
Avishai Weizman*\\
\textit{School of Electrical and Computer Engineering}\\
\textit{Ben-Gurion University of the Negev}\\
Beersheba, Israel\\
Email: wavishay@post.bgu.ac.il\\
ORCID: 0009-0004-1182-8601
\end{minipage}

}\\[0.9em]
\makebox[\textwidth][c]{
\begin{minipage}[t]{0.33\textwidth}\centering
Liav Cohen\\
\textit{Intelligence Systems}\\
\textit{Afeka College of Engineering}\\
Tel Aviv, Israel\\
Email: liav.cohen@s.afeka.ac.il\\
ORCID: 0009-0009-2756-8783
\end{minipage}%

\begin{minipage}[t]{0.33\textwidth}\centering
Moshe Tshuva\\
\textit{Mechanical Engineering}\\
\textit{Afeka College of Engineering}\\
Tel Aviv, Israel\\
Email: moshet@afeka.ac.il\\
ORCID: 0000-0002-0828-5595
\end{minipage}%
\hfill
\begin{minipage}[t]{0.33\textwidth}\centering
Yehudit Aperstein\\
\textit{Intelligence Systems}\\
\textit{Afeka College of Engineering}\\
Tel Aviv, Israel\\
Email: apersteiny@afeka.ac.il\\
ORCID: 0000-0002-0828-5595
\end{minipage}%
}%
}

\maketitle

\begin{abstract}
We address the challenge of detecting rare and diverse anomalies in surveillance videos using only video-level supervision. Our dual-backbone framework combines convolutional and transformer representations through top-$k$ pooling, achieving 90.7\% area under the curve (AUC) on the UCF-Crime dataset.
\end{abstract}


\section{Introduction}
Automatically detecting abnormal events in surveillance
videos remains a challenge because such events are rare. 
Obtaining precise temporal annotations for a large-scale dataset
is often impractical due to the high cost of labeling, ambiguity
in defining anomaly boundaries, and inconsistencies between
annotators. These constraints make \textit{weakly supervised learning}
with video-level labels an efficient alternative for real-world
settings.
\textit{Multiple Instance Learning} (MIL) supports this setting by treating each video as a bag of temporal segments, where an anomalous label implies that at least one segment is abnormal.
Early approaches~\cite{sultani2018real} and later frameworks such
as MSTAgent-VAD~\cite{zhao2025mstagent} employ
MIL with video-level labels to localize anomalies, incorporating temporal smoothness and attention mechanisms. In~\cite{tian2021weakly}, this paradigm was further improved through \textit{Robust Temporal Feature Magnitude} (RTFM) learning, which enhances MIL by separating feature magnitudes and incorporating multi-scale temporal modeling to reveal informative abnormal cues.
Other perspectives use \textit{multi-sequence learning}~\cite{li2022selftraining} enhances temporal consistency by ranking short sequences of consecutive snippets and refining their labels through a transformer-based self-training process, while the clustering method in~\cite{zaheer2020clustering} improves stability by encouraging compact grouping of normal patterns and increasing their separation from anomalous ones.

In~\cite{strijbosch2024usable} proposed a class-agnostic anomaly detection pipeline that leverages \textit{Inflated 3D ConvNet} (I3D) feature embeddings and threshold-based scoring, providing a simpler but less discriminative baseline for surveillance data. 

In this work, we propose a dual-backbone MIL network that combines two complementary video encoders: an I3D model~\cite{carreira2017quo} for spatiotemporal motion analysis and a transformer-based architecture (TimeSformer)~\cite{bertasius2021TimeSformer} to enrich the overall video representation. The fused outputs of these encoders are processed by lightweight \textit{fully connected} (FC) layers, and the resulting features are aggregated using top-$k$ pooling as used in~\cite{wu2023topk} to produce video-level predictions.

\section{Proposed Method}
The proposed method is suitable for the UCF-Crime dataset~\cite{sultani2018real}, where only about $6\%$ of the video durations are longer than ten minutes, while most video durations last only a few minutes.
Following the approach of~\cite{sultani2018real} and~\cite{zhao2025mstagent}, our process begins by considering a collection of videos defined as
$V = \{v_1, v_2, \dots, v_M\}$, where $M$ represents the total number of videos in the dataset. 
Each video $v_m$ is divided into a sequence of segments
$v_m = \{u_{m,1}, u_{m,2}, \dots, u_{m,N}\}$, where $N$ denotes the total number of segments in video $v_m$. These segments serve as the fundamental processing units for our encoders. Every video $v_m$ is 
associated with a single video-level label $y_m \in \{0,1\}$, 
indicating whether the video is normal or anomalous.
Each video is uniformly divided into $N = 32$ temporal segments. From each segment $u_{m,i}$, 16 frames are uniformly sampled, forming a shorter segment $x_{m,i} = \{F_{m,i}^{1}, F_{m,i}^{2}, \dots, F_{m,i}^{16}\}$. Segments containing fewer than 16 frames are padded by repeating the last frame until reaching the fixed length of 16 frames. This sampling strategy provides an effective balance between computational cost and temporal coverage, and is well-suited to the UCF-Crime dataset~\cite{sultani2018real}, where long segments are relatively rare. Each shorter segment is then encoded using two pretrained backbones: the I3D encoder (768-dimensional vector) and the TimeSformer encoder (1024-dimensional vector), both operating on 16-frame inputs. The extracted features are $\ell_2$-normalized and concatenated to form a unified feature representation (1792-dimensional vector).
The objective is to learn a function $f(x_{m,i}; \theta)$ that assigns higher anomaly scores to abnormal segments while producing lower scores for normal ones. Each segment feature $f_{m,i}$, derived from the corresponding segment $x_{m,i}$, is then passed through four FC layers that output a scalar anomaly score. A LeakyReLU activation is applied between layers, producing a final score $s_i$ that indicates the degree of abnormality for the $i$-th segment. Following the MIL assumption, only some segments in an anomalous video are abnormal, whereas all segments in a normal video are regular. To obtain a video-level representation, the $k$ highest segment scores are aggregated through top-$k$ pooling, formulated as:

\begin{equation}
z(V) = \frac{1}{k}\sum_{i \in \mathrm{TopK}(s,k)} s_i,
\end{equation}

where $z(V)$ denotes the video-level anomaly logit, $\hat{y} = \sigma(z(V))$ represents the predicted anomaly probability obtained through the sigmoid function $\sigma(\cdot)$, and $k$ is a hyper-parameter that determines the number of top segments used in the aggregation.
All components of the model are jointly optimized using the binary cross-entropy loss.

\begin{figure*}[t]
    \centering
      \includegraphics[width=0.90\textwidth, height=0.20\textheight, keepaspectratio]{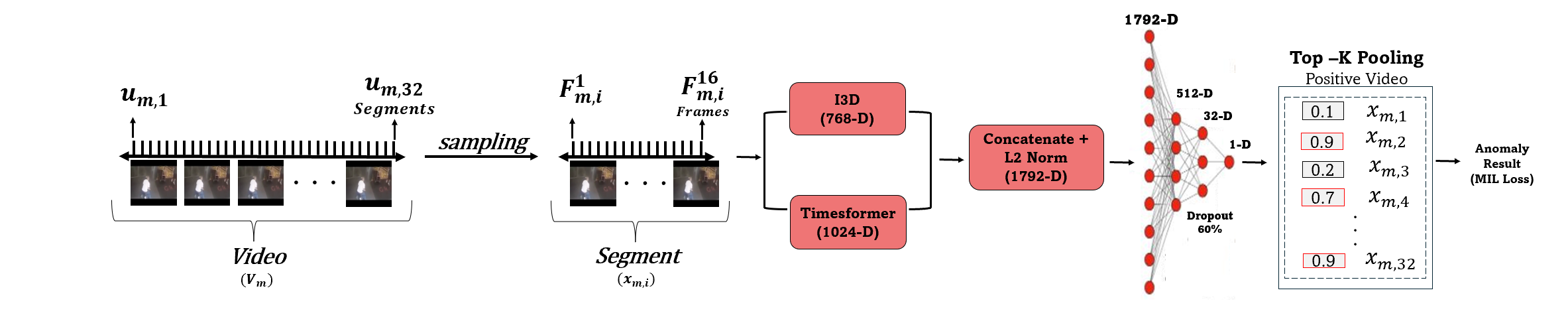}
    \caption{Illustration of the dual-backbone MIL framework. 
    Each video is divided into 32 temporal segments ($v_m$). From each segment ($u_{m,i}$), 16 frames ($x_{m,i}$) 
    are uniformly sampled to form a shorter segment, which is encoded by I3D (convolutional-based) and TimeSformer  (transformer-based) encoders. 
    The concatenated and $\ell_2$-normalized features are processed by a compact prediction head 
    and aggregated through top-$k$ pooling to produce the final video-level anomaly prediction.}
    \label{fig:pipeline}

\end{figure*}

The proposed architecture, illustrated in~\autoref{fig:pipeline}, computes segment-level anomaly scores, then applies top-$k$ pooling to produce the final video prediction.

\section{Experiments and Results}\label{sec:Experiments_and_results}
Anomalies in the UCF-Crime dataset~\cite{sultani2018real} are defined as irregular or criminal activities captured in untrimmed surveillance videos covering $13$ anomaly categories and normal events. Each video is annotated at the video level, with varying durations, scenes, and complexity.
\begin{table}[t]
\centering
\caption{Comparison of anomaly detection methods on the UCF-Crime dataset.}
\label{tab:ucf_comparison}
\begin{tabular}{|l|l|c|}
    \hline
    \textbf{Method} & \textbf{Model} & \textbf{AUC (\%)} \\ \hline
    \textbf{Ours} & Feature fusion + Top-$k$ MIL & \textbf{90.7} \\ \hline
    Zhao et al.~\cite{zhao2025mstagent} & Transformer (VideoSwin) & 89.3 \\ \hline
    Wu et al.~\cite{wu2024weakly} & CLIP + Prompt Learning & 88.4 \\ \hline
    Li et al.~\cite{li2022selftraining} & Multi-Sequence Learning (VideoSwin) & 85.6 \\ \hline
    Tian et al.~\cite{tian2021weakly} & I3D & 84.3 \\ \hline
    CLAWS Net+~\cite{zaheer2020clustering} & Clustering-Aided MIL & 84.1 \\ \hline
    Zhong et al.~\cite{zhong2019gcn} & GCN-LNC (GCN + Classifier) & 82.1 \\ \hline
    Strijbosch~\cite{strijbosch2024usable} & I3D + Clustering & 80.7 \\ \hline
    Sultani et al.~\cite{sultani2018real} & C3D + MIL-ranking & 75.4 \\ \hline
    Cho et al.~\cite{cho2022unsupervised} & ITAE + Normalizing Flows & 70.9 \\ \hline
\end{tabular}
\end{table}
As shown in~\autoref{tab:ucf_comparison}, our novel dual-backbone fusion model achieves the highest AUC, outperforming all competing approaches. The compared approaches span a wide range of strategies to tackle this task, including transformer-based models, \textit{contrastive language-image pre-training} (CLIP)~\cite{Radford2021LearningTV}-based vision-language methods, I3D combined with clustering, and graph-based approaches such as \textit{graph convolutional networks} (GCNs). Although these approaches tackle anomaly detection from different angles, our dual-encoder model achieves superior AUC performance over all other methods on the UCF-Crime dataset.

\section{Conclusions and future work}
\label{sec:Conclusions_and_future_work}
This work presented a novel architecture for video anomaly detection that integrates I3D and TimeSformer encoders with lightweight FC layers.
The proposed framework demonstrated strong performance on the UCF-Crime dataset, achieving a high AUC despite being restricted to a uniform frame-sampling strategy due to computational limitations.
This design was guided by the assumption that the UCF-Crime dataset rarely contains long continuous videos, making uniform sampling sufficient to capture the relevant temporal dynamics.
In future work, we plan to extend the framework toward multi-class anomaly detection and investigate advanced sampling strategies to enhance the temporal representation of the video segments.

\bibliographystyle{IEEEtran}
\bibliography{references} 

@inproceedings{sultani2018real,
  title     = {Real-world anomaly detection in surveillance videos},
  author    = {Sultani, Waqas and Chen, Chen and Shah, Mubarak},
  booktitle = {Proceedings of the IEEE Conference on Computer Vision and Pattern Recognition (CVPR)},
  pages     = {6479--6488},
  year      = {2018}
}

@article{zhao2025mstagent,
  title     = {MSTAgent-VAD: Multi-scale Video Anomaly Detection Using Time Agent Mechanism for Segments' Temporal Context Mining},
  author    = {Zhao, Shuang and Zhao, Rui and Meng, Yifeng and Gu, Xiaoqing and Shi, Chunlei and Li, Dong},
  journal   = {Expert Systems with Applications},
  volume    = {276},
  pages     = {127154},
  year      = {2025},
  publisher = {Elsevier}
}

@inproceedings{tian2021weakly,
  title     = {Weakly-supervised Video Anomaly Detection with Robust Temporal Feature Magnitude Learning},
  author    = {Tian, Yu and Pang, Guansong and Chen, Yuanhong and Singh, Rajvinder and Verjans, Johan W. and Carneiro, Gustavo},
  booktitle = {Proceedings of the IEEE/CVF International Conference on Computer Vision (ICCV)},
  pages     = {4975--4986},
  year      = {2021}
}

@inproceedings{Radford2021LearningTV,
  title={Learning Transferable Visual Models From Natural Language Supervision},
  author={Alec Radford and Jong Wook Kim and Chris Hallacy and Aditya Ramesh and Gabriel Goh and Sandhini Agarwal and Girish Sastry and Amanda Askell and Pamela Mishkin and Jack Clark and Gretchen Krueger and Ilya Sutskever},
  booktitle={International Conference on Machine Learning},
  year={2021},
}

@inproceedings{zhong2019gcn,
  title     = {Graph Convolutional Label Noise Cleaner: Train a Plug-and-Play Action Classifier for Anomaly Detection},
  author    = {Zhong, Jiang and Li, Shuai and Kong, Tao and Liu, Changsong and Xie, Yuning and Li, Qian and Ji, Rongrong},
  booktitle = {Proceedings of the IEEE/CVF Conference on Computer Vision and Pattern Recognition (CVPR)},
  pages     = {1237--1246},
  year      = {2019}
}

@inproceedings{zaheer2020clustering,
  title     = {Clustering Assisted Weakly Supervised Learning with Normalcy Suppression for Anomalous Event Detection},
  author    = {Zaheer, Muhammad Zaigham and Mahmood, Arif and Astrid, Marcella and Lee, Seung-Ik},
  booktitle = {Computer Vision -- ECCV 2020},
  series    = {Lecture Notes in Computer Science},
  volume    = {12367},
  pages     = {358--376},
  year      = {2020},
  publisher = {Springer}
}

@inproceedings{li2022selftraining,
  title     = {Self-Training Multi-Sequence Learning with Transformer for Weakly Supervised Video Anomaly Detection},
  author    = {Li, Shuo and Liu, Fang and Jiao, Licheng},
  booktitle = {Proceedings of the AAAI Conference on Artificial Intelligence (AAAI)},
  volume    = {36},
  number    = {2},
  pages     = {1395--1403},
  year      = {2022}
}

@mastersthesis{strijbosch2024usable,
  title      = {Towards a Usable Crime-Based Anomaly Detection Model},
  author     = {Strijbosch, Daan},
  school     = {Delft University of Technology},
  year       = {2024}
}

@article{cho2022unsupervised,
  title   = {Unsupervised Video Anomaly Detection via Normalizing Flows with Implicit Latent Features},
  author  = {Cho, MyeongAh and Kim, Taeoh and Kim, Woo Jin and Cho, Suhwan and Lee, Sangyoun},
  journal = {Pattern Recognition},
  volume  = {129},
  pages   = {108703},
  year    = {2022}
}

@inproceedings{wu2024weakly,
  title     = {Weakly Supervised Video Anomaly Detection and Localization with Spatio-Temporal Prompts},
  author    = {Wu, Peng and Zhou, Xuerong and Pang, Guansong and Yang, Zhiwei and Yan, Qingsen and Wang, Peng and Zhang, Yanning},
  booktitle = {arXiv preprint arXiv:2408.05905},
  year      = {2024}
}

@inproceedings{carreira2017quo,
  title     = {Quo Vadis, Action Recognition? A New Model and the Kinetics Dataset},
  author    = {Carreira, Joao and Zisserman, Andrew},
  booktitle = {Proceedings of the IEEE Conference on Computer Vision and Pattern Recognition (CVPR)},
  pages     = {6299--6308},
  year      = {2017}
}

@inproceedings{bertasius2021timesformer,
  title     = {Is Space-Time Attention All You Need for Video Understanding?},
  author    = {Bertasius, Gedas and Wang, Heng and Torresani, Lorenzo},
  booktitle = {Proceedings of the IEEE/CVF Conference on Computer Vision and Pattern Recognition (CVPR)},
  pages     = {6689--6699},
  year      = {2021}
}

@inproceedings{wu2023topk,
  title     = {Top-K Pooling with Patch Contrastive Learning for Weakly-Supervised Semantic Segmentation},
  author    = {Wu, Wangyu and Dai, Tianhong and Huang, Xiaowei and Ma, Fei and Xiao, Jimin},
  booktitle = {Proceedings of the IEEE International Conference on Systems, Man, and Cybernetics (SMC)},
  pages     = {5270--5275},
  year      = {2024}
}

\end{document}